# Development and Feasibility Evaluation of an Edge AI as Medical Device System for Breast Cancer Multidisciplinary Team Meetings

**Authors:**

Aarzoo Dhiman[1] (A.Dhiman@hull.ac.uk), Farzana Haque[2] (Farzana.Haque4@nhs.net), Iqtedar Muazzam[2] (iqtedar.muazzam@nhs.net), Kartikae Grover[2] (kartikaegrover@nhs.net), Lydia Brian Smith[3] (L.Bryan-Smith-2014@hull.ac.uk), William Stephen Jones[1] (Will.Jones@hull.ac.uk)

**Affiliations:**

[1] Centre of Excellence for Data Science, Artificial Intelligence and Modelling (DAIM), Faculty of Science and Engineering, University of Hull, Hull, United Kingdom

[2] Queen's Centre for Oncology and Haematology, Hull University Teaching Hospitals NHS Foundation Trust, Castle Hill Hospital, Cottingham, United Kingdom

[3] Department of Computer Science, Faculty of Science and Engineering, University of Hull, Hull, United Kingdom

**Corresponding author:** Aarzoo Dhiman, Centre of Excellence for Data Science, Artificial Intelligence and Modelling (DAIM), Faculty of Science and Engineering, University of Hull, Hull, United Kingdom. Email: a.dhiman@hull.ac.uk

# Abstract

Cancer Multidisciplinary Team (MDT) meetings manage a high number of complex cases under considerable time pressure, while increased documentation requirements can reduce clinical efficiency and decision quality. We developed a fully on-device Retrieval-Augmented Generation (RAG)-based Large Language Model (LLM) pipeline relying solely on open-source models. The primary contributions include transcription of breast cancer MDT discussions, National Institute for Health and Care Excellence (NICE) guideline-grounded treatment recommendation generation, and clinical evaluation of the transcriptions and generated treatments. The system was evaluated using two recorded simulated MDT discussions, ten clinically validated synthetic discussions, and 1,270 acoustically augmented variants. The open-source Automatic Speech Recognition (ASR) model, Whisper Large v3, was refined using silence-aware preprocessing and noise-adaptive decoding, achieving performance within 0.58 percentage points in word error rate and 1.58 percentage points in word information lost of a commercial clinical ASR benchmark on augmented audio. The local MedGemma-RAG model identified 2.3 times more MDT-concordant interventions than a proprietary comparator (recall 0.318 vs 0.136; difference 0.18, 95% CI 0.041–0.327; p=0.020), with no significant difference in overall accuracy. The proposed intervention-based prompt engineering facilitated evaluation while improving generation of clinically relevant recommendations. Stakeholders identified automated documentation, treatment recommendation support, and case triage as promising applications, while highlighting workflow integration, governance, and clinician trust as implementation barriers. The use of open-source models on a single NVIDIA Jetson AGX Orin device, together with comparison with proprietary models, demonstrates the potential of an Edge-AI approach to provide privacy-preserving clinical support without reliance on paid cloud-based models.

**Keywords:** clinical decision support; multidisciplinary team meeting; large language models; retrieval-augmented generation; edge computing; AI as a medical device

# 1. Introduction

Breast cancer is the most frequently diagnosed cancer in the United Kingdom, with over 60,000 new cases annually [1, 2]. Management of breast cancer is inherently complex and increasingly multidisciplinary. Treatment decisions must integrate radiological findings, histopathology, tumour biology, disease stage, comorbidities, prior treatments, and patient preferences, alongside eligibility for evolving systemic therapies and clinical trials [1, 3]. Multidisciplinary team (MDT) meetings, or tumour boards, are the mechanism through which this integration occurs and remains the standard for complex cancer care delivery [1, 3-5].

Despite their central role, MDT meetings face significant operational and cognitive challenges [1, 3, 6]. Cancer MDTs review large and growing caseloads with only a few minutes per patient [7], and prolonged sequential decision-making is associated with measurable decision fatigue, reduced information quality and diminished team contribution across a meeting [6]. Missing results and incomplete records compound this, and national guidance has called for MDT processes to be streamlined [8, 9]. Consequently, a widening gap exists between the volume of clinically relevant information and what can realistically be synthesised during the meeting.

Particularly in breast oncology, this gap is particularly pronounced, where treatment recommendations depend on detailed combinations of clinicopathological features and the relevant guidance is distributed across multiple sources, including National Institute for Health and Care Excellence (NICE guidelines), technology appraisals, and international consensus statements [10, 11].

In response, there has been growing interest in digital tools to support MDT processes, including case preparation, presentation, documentation, and decision-making [1, 12]. More recently, the interest in Artificial Intelligence (AI) for oncology decision support has expanded rapidly [13, 14]. In the UK, AI software with a medical purpose is regulated as AI as a medical device (AIaMD) within the software as a medical device framework [15, 16]. Recent developments in AI have introduced complementary technologies that address different stages of the MDT workflow. Automatic speech recognition (ASR) captures discussion in real time, supporting documentation, traceability and audit [17, 18], although performance deteriorates with background noise, overlapping speech, and specialist terminology [19-21]. Large language models (LLMs) can summarise and structure clinical information, with tumour board studies reporting moderate-to-high but variable concordance with MDT decisions; however, they perform less consistently for treatment sequencing and complex therapeutic recommendations, underscoring the need for expert oversight [22-26]. Retrieval-augmented generation (RAG) grounds outputs in retrieved documents at inference, improving factual accuracy and guideline alignment while reducing unsupported responses [27-29]. A rapid scoping review of LLM, RAG and AI Decision Support in Breast Cancer MDT Settings is provided in Supplementary Table 1.

Nevertheless, important limitations remain in the current literature. First, existing MDT decision-support studies predominantly evaluate structured referral letters or curated case summaries rather than live MDT discussions and therefore do not address the conversational and unstructured nature of tumour board meetings; the closest RAG work for breast tumour boards similarly uses curated case text as input [30, 31]. Consequently, they neither address the challenges of speech-derived clinical information nor evaluate end-to-end workflows from discussion to recommendation. Second, and more consequentially for adoption, most reported systems rely on proprietary cloud-hosted models. Beyond governance, confidentiality, cost and reproducibility concerns, reliance on external inference increases the risk of information leakage and presents a major barrier to deployment in healthcare systems such as the NHS [32].

This study addresses these limitations by developing and evaluating an AIaMD pipeline built using open-source ASR and LLM models, that captures MDT discussions, transcribes and structures clinical information, and generates NICE guideline-informed recommendations from MDT conversations. The entire pipeline runs on a single edge device using quantised models, enabling local inference without external data transmission. We introduce preprocessing and post-processing steps for a general-purpose ASR model to improve transcription quality and achieve performance comparable to that of proprietary clinical ASR systems. We further investigate intervention-based prompting strategies to direct RAG-LLM outputs towards predefined clinically relevant treatment interventions while allowing uncertainty when information is insufficient. We further propose two expert-informed evaluation approach that distinguishes clinically appropriate additional interventions from true errors, enabling a more clinically meaningful assessment of LLM-generated treatment recommendations.

# 2. Methods

## 2.1. Study Design

This development and bench (non-clinical) feasibility study evaluated an edge-deployed AIaMD pipeline integrating ASR, a locally hosted LLM, and RAG. No patients were recruited, and no identifiable patient data were processed. Evaluation comprised of component-level assessment of transcription accuracy, clinical terminology preservation, and recommendation quality, together with system-level assessment of clinical relevance and concordance with MDT recommendations. At the system level, outputs were assessed for their clinical relevance and alignment with MDT decision-making, providing an initial indication of potential utility as a potential AIaMD decision-support tool.

## 2.2. Edge Deployment Architecture

The pipeline was deployed on an NVIDIA Jetson AGX Orin Developer Kit (64 GB) [33] with a 4 TB external solid-state drive for model storage, preserving unified memory for inference. The software stack comprised Whisper large-v3 [34], nomic-embed-text [35], FAISS [36], and quantised GGUF models (27B–70B), orchestrated using CUDA, Docker, TensorRT, and llama.cpp. The 64 GB unified memory enabled local inference with high-parameter models that would otherwise require cloud-based GPU infrastructure. All processing was performed on device; audio, transcripts, embeddings, and model outputs remained within local hardware and institutional infrastructure, supporting data locality and compatibility with NHS information governance requirements.

## 2.3. Data Sources

Three datasets were used and are reported separately because they provide different levels of evidence.

*Tier 1: Authentic simulated MDT recordings.* Members of the Hull University Teaching Hospitals (HUTH) NHS breast MDT, including surgeons, oncologists, radiologists, pathologists, specialist nurses, and MDT coordinators, recorded discussions of previously encountered cases while performing their usual MDT roles. All recordings were fully anonymised, preserving authentic clinical terminology, conversational dynamics, and decision-making without providing identifiable patient information. Two recordings were obtained, *Recording 1*: 7 min 44 s and *Recording 2*: 10 min 20 s after removal of introductory discussion. Audio was captured and stored on a University of Hull-managed device without any data transfer to clouds. A ground truth transcription was generated and subsequently reviewed and verified by a consultant oncologist (FH, co-author) to ensure accuracy of clinical terminology and fidelity to the intended medical context.

*Tier 2: Synthetic MDT discussions.* Ten multi-speaker MDT scripts were generated from a consultant-authored template using a general-purpose LLM, reviewed for clinical plausibility by FH, and synthesised using Amazon Polly (16 kHz, 16-bit, mono) [37] with role-specific voices and accents (Supplementary Table 2). Fig. 1 provide the workflow for generation of synthetic dataset, with the detailed characteristics of the synthetic data provided in Supplementary Table 3.

*Tier 3: Acoustically augmented audio.* Each synthetic (Tier 2) recording was augmented using all combinations of background noise, room impulse response convolution, speech overlap, speed perturbation, pitch shifting, and signal dropout, generating 127 variants per recording and a total of 1,270 augmented audio files. The specific details, including implementation methods and parameter ranges used to generate the augmented evaluation dataset have been provided in the Supplementary Table 4. The workflow used to generate the Tier 2 and Tier 3 datasets is shown in Fig. 1.

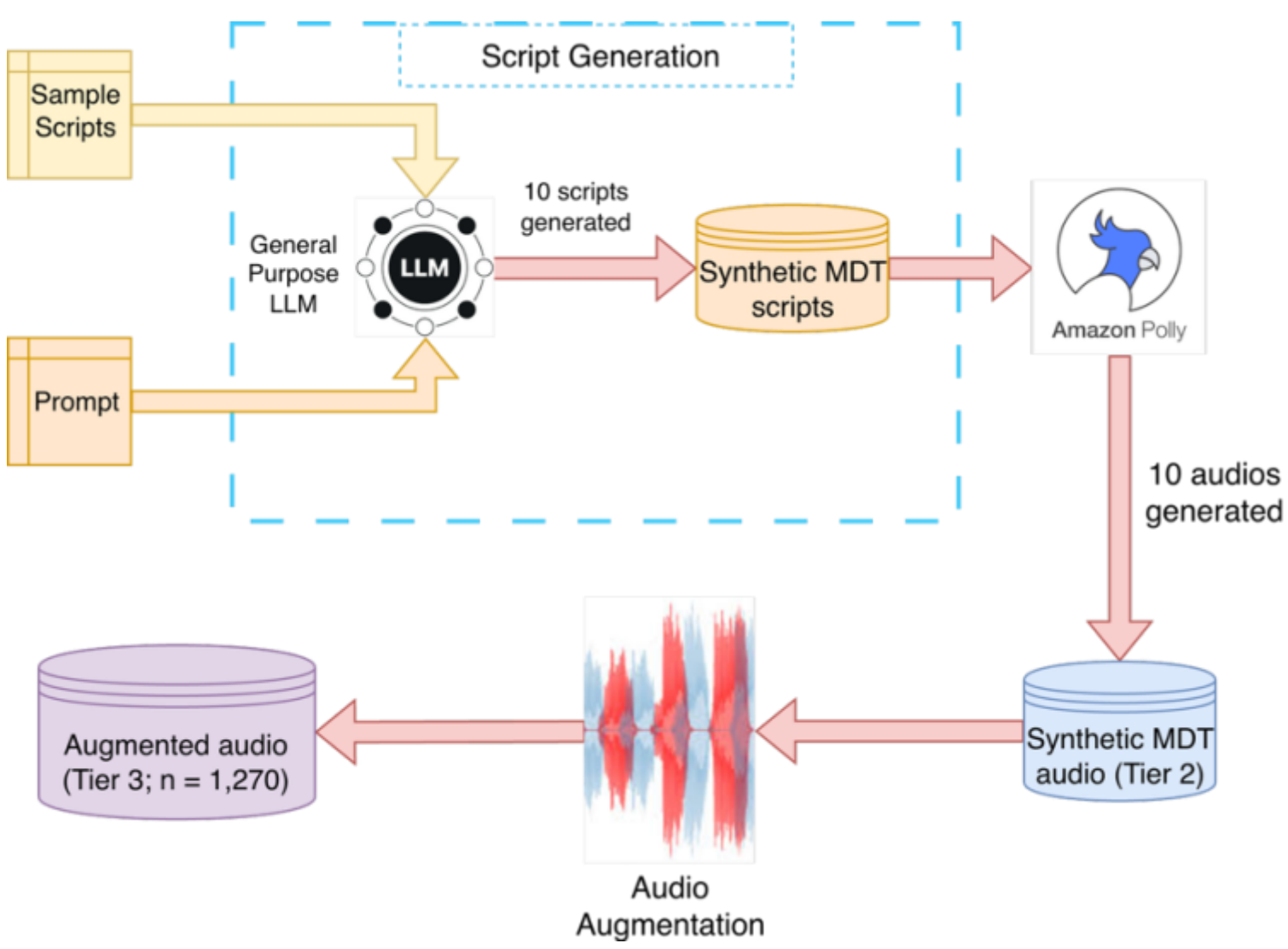

**Fig. 1** Workflow for generation of synthetic and augmented datasets. Sample MDT scripts and a prompt template were used to generate synthetic MDT discussions with a general-purpose LLM. Discussions were synthesised using Amazon Polly to produce Tier 2 audio, which was subsequently augmented using acoustic transformations to generate the Tier 3 evaluation dataset.

The reference transcript (ground truth) for Tier 1 comprised oncologist-verified transcripts, whereas the original LLM-generated synthetic MDT scripts were used for Tiers 2 and 3. As the latter audio files were derived from text-to-speech synthesis, they did not reproduce spontaneous speech characteristics such as disfluencies, interruptions, false starts, or variations in microphone distance. These datasets were therefore used for controlled development and benchmarking, while the Tier 1 recordings were reserved as a held-out evaluation set to assess performance on representative MDT conversations.

## 2.4 Transcription Pipeline and Evaluation

Four ASR systems were evaluated in their default configurations: (i) Whisper large-v3 [34], (ii) NVIDIA Parakeet [38], and (iii) WhisperX [39], all open-source and locally deployable, aligning with the objective of developing a fully edge-deployable pipeline and (iv) Amazon Transcribe Medical (AWS) [40], which was included only as a commercial clinical benchmark, representing an industry-standard system optimised for clinical speech recognition. All evaluated models are primarily trained on American English datasets, which may influence performance in MDT environments characterised by diverse accents and speech patterns. A brief description of the models has been provided in Supplementary Table 5.

Transcription quality was assessed using Word Error Rate (WER) and Word Information Lost (WIL) following normalisation with the Whisper English text normaliser. WER quantifies word-level errors by measuring insertions, deletions, and substitutions, thereby estimating the edit cost of a transcription. In contrast, WIL measures information loss by accounting for both missed and additional words in a more proportionate manner, and is therefore useful for assessing the potential impact of transcription errors on the information content of clinical speech [41] (See Supplementary Section 3.1). Both WIL and WER assign equal importance to each word, which limits their suitability in clinical contexts. In medical transcription, misrecognizing function words (e.g., transcribing "the" as "a") represents a minor error with negligible semantic impact, whereas confusing clinically distinct terms such as "hypothyroidism" and "hyperthyroidism" may result in significantly altered interpretation and potentially life-threatening clinical decisions. Therefore, we also employ domain-specific medical term–based evaluation criteria to assess the performance of the models [42, 43].

To assess domain-specific medical term fidelity, we extracted a curated set of clinically relevant phrases from the ground-truth MDT transcripts, including breast cancer-specific diagnostic and treatment terminology (see Supplementary Section 3.2). For each transcription, the text was tokenised and n-grams of length 1–6 were generated and compared with the reference phrases using exact matching and fuzzy matching (Levenshtein similarity ≥80%, implemented using FuzzyWuzzy). Recall was used to quantify medical term recognition performance. Precision was not calculated because candidate terms were restricted to the predefined vocabulary, limiting the relevance of false-positive detection in this evaluation.

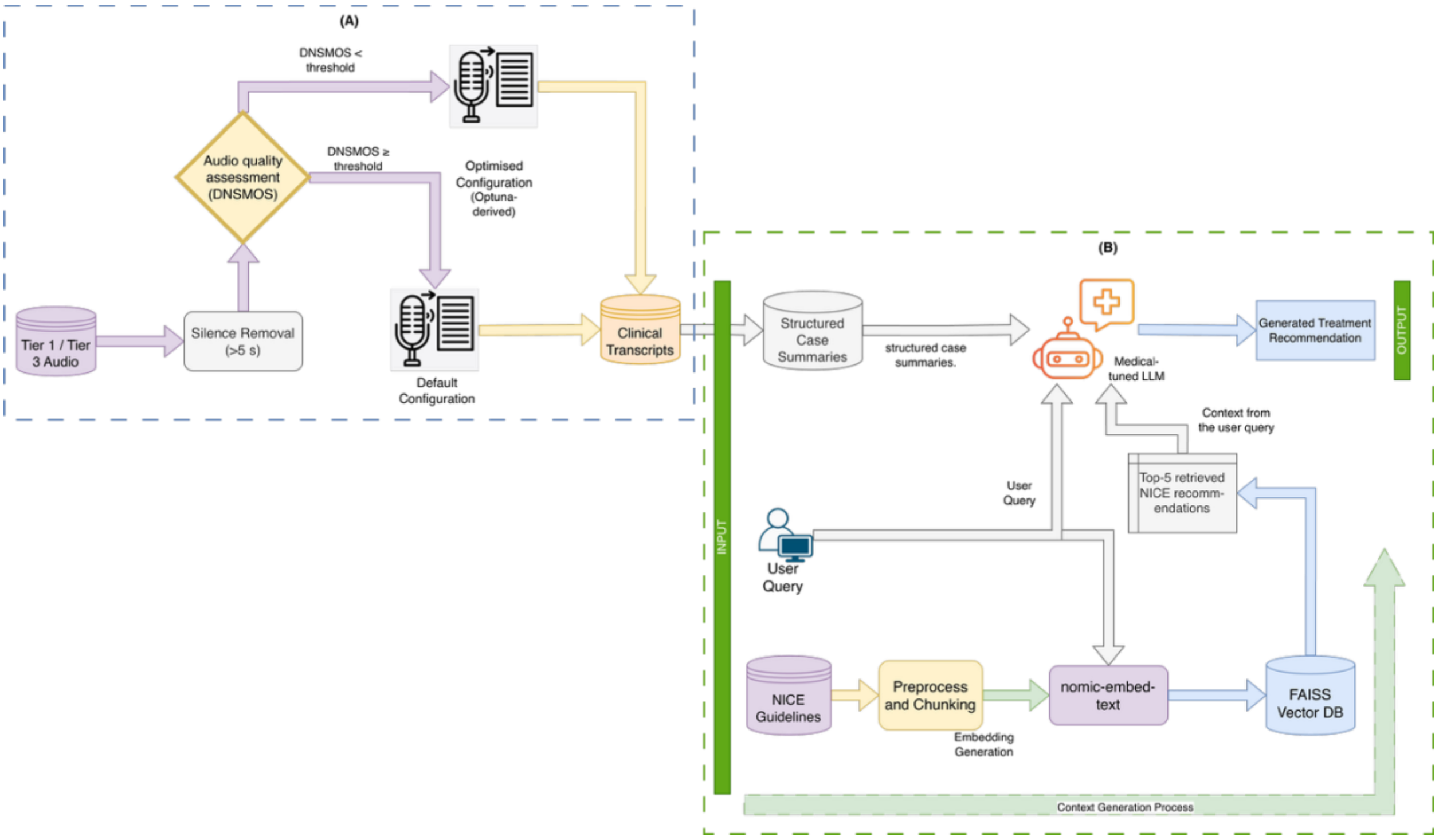

**Fig. 2** Overview of the proposed edge-deployed AI pipeline. (A) Adaptive transcription pipeline. Audio recordings (Tier 1 and Tier 3) undergo silence removal followed by DNSMOS-based quality assessment. Depending on the estimated audio quality, recordings are transcribed using either the default or an optimised Whisper configuration to produce clinical transcripts. (B) Guideline-informed recommendation pipeline. Clinical transcripts are converted into structured case summaries, which are embedded and used to retrieve the five most relevant NICE guideline recommendations from a FAISS vector database. Retrieved guideline context and the structured case summary are provided to MedGemma to generate treatment recommendations.

The best-performing model was selected based on standard transcription accuracy (WER and WIL), clinical term recall, and compatibility with local edge-based deployment. Although domain-specific fine-tuning is a potential approach to improve ASR performance, it typically requires substantial in-domain speech or transcript data and additional computational resources for model adaptation [44]. For example, recent clinical applications of Whisper fine-tuning have used approximately 1,300 hours of clinical speech, illustrating the data requirements of full domain adaptation [45]. In contrast, our aim was to achieve targeted improvements using the limited data and computational resources available for this study, while retaining the advantages of a general-purpose open-source model. This approach was selected to achieve practical improvements with the limited data and computational resources available for the study. The selected open-source ASR model was subsequently optimised using a three-stage pipeline (Fig. 2 (A)) to address repetition artefacts and hallucinations associated with silence and background noise. First, to reduce hallucinations during silent periods, audio was preprocessed using the FFmpeg silencedetect filter (−23 dB threshold). Silent segments longer than 5 s were removed while retaining 2 s of surrounding silence before concatenating the remaining audio.

The ASR model hyperparameters were then optimised using Bayesian optimisation with Optuna [46]. A total of 100 trials were performed using parallel search, with WER minimisation as the objective on Tier 2 and Tier 3 audio. WER was selected as the optimisation objective because it provides a direct and stable word-level error signal for hyperparameter optimisation, while WIL was retained as a complementary metric to assess information loss. As the Tier 2 and 3 audio contained substantial noise, the optimised hyperparameters were sensitive to acoustic conditions. We therefore implemented a noise-adaptive inference strategy in which the audio noise level was quantified using the Deep Noise Suppression Mean Opinion Score (DNSMOS) background-noise score [47], ranging from 1 (very noisy) to 5 (clean). Default or optimised ASR parameters were applied according to the estimated noise level.

Hyperparameter optimisation and threshold selection were performed exclusively on the Tier 2 and Tier 3 augmented dataset, with DNSMOS thresholds evaluated from 2.0 to 5.0 in 0.5 increments. The resulting strategy was subsequently evaluated on the held-out Tier 1 recordings, with results presented in Section 3.

## 2.5 LLM Processing and Recommendation Generation

Treatment prediction comprised three stages: extraction of structured case information from the MDT transcript, treatment recommendation generation without guideline retrieval, and recommendation generation with retrieval-augmented generation (RAG). Two domain-adapted LLMs were evaluated for case information extraction: MedGemma 27B (text-only, 8-bit quantised GGUF, ~31.8 GB, 128k context) [48] and Palmyra-Med 70B (3-bit quantised IQ3_M GGUF, ~31.9 GB, 32k context) [49], with decoding parameters provided in Supplementary Table 8. These models were selected due to their training on biomedical corpora and their ability to process long clinical inputs via extended context windows.

### 2.5.1 Case Description and Treatment Separation

Extraction of individual clinical case descriptions is important for the end-to-end pipeline because the audio transcription is continuous and does not contain explicit delimiters between separate cases. Errors in case separation may propagate to subsequent treatment-plan extraction and recommendation generation. A one-shot prompting approach has proven to yield better results for similar tasks [50], was therefore used to correct transcription artefacts, standardise medical terminology, and extract (i) structured case descriptions and (ii) corresponding treatment plans without introducing new clinical information. We deliberately use the term "case description" rather than "summary" to distinguish this task from the summarisation capabilities of LLMs. The system prompts are provided in Supplementary Section 5.1.

This process was evaluated using the Tier 1 audio data, as Tiers 2 and 3 were primarily used to assess model performance under controlled and noisy conditions. Tier 1 contained six clinical cases. We manually reviewed the extracted case boundaries to assess the quality of case separation; however, this assessment was not quantitatively measured and should therefore be considered a limitation of the study.

### 2.5.2 Treatment Prediction Without Guideline Context

The second stage evaluated intrinsic clinical reasoning by generating treatment recommendations without access to external guideline content. Only the extracted case description was provided as input. The treatment portion of the transcript was excluded to prevent information leakage. Although prompts instructed models to follow NICE breast cancer guidelines (detailed shortly), no guideline text was provided. This setup evaluates whether guideline knowledge is implicitly encoded within the models. Prompts used are provided in Supplementary Section 5.2. We manually reviewed the generated outputs from this step.

### 2.5.3 Treatment Prediction with Guideline Context (RAG)

The RAG knowledge base was constructed from NICE guidelines CG81 [51], CG164 [52], and NG101 [53] (total 177 pages). The LLM and retrieval-augmented generation workflow is illustrated in Fig. 2 (B). Guidelines were segmented into individual recommendation-level chunks while preserving section headings, recommendation identifiers, source guideline, and page number as metadata. Tables were extracted using Camelot [54] and stored in the same format. Chunks were embedded using nomic-embed-text [35] and indexed in FAISS [36]. During inference, each case summary was embedded, and the five most relevant guideline chunks were retrieved and incorporated into the generation prompt.

RAG-based treatment recommendations were evaluated using two output formats: *free-text recommendation generation* and *structured intervention prediction*. We evaluated free-text recommendations using two prompt variants to assess the effect of prompt formulation. The first prompt generated treatment recommendations directly from the case description (FT1, Supplementary Section 5.3.1), requiring the model to map the available clinical information to a NICE-consistent treatment plan. In contrast, FT2 (Supplementary Section 5.3.2) first required the model to reconstruct the MDT's reasoning, decision drivers, conditional logic, treatment sequencing, referrals, and biomarker-dependent branches before generating the treatment plan.

Although free-text outputs provide a natural representation of clinical recommendations, their interpretation and evaluation require review of the complete generated text and are less amenable to *direct* calculation of precision and recall. We therefore introduced a structured intervention format in which the model classified predefined treatment interventions (Supplementary Section 5.4) as *Yes*, *No*, or *Not Enough Information (NEI)*. This approach reduced the burden of evaluation by allowing assessors to define the reference interventions once for each case, rather than reviewing every generated response individually. For the six cases and two models evaluated using the two free-text prompting strategies, this required review of 24 generated outputs, whereas the structured approach required six case-level reference annotations. The structured format also enabled direct calculation of precision and recall and provided the model with a predefined set of clinically relevant interventions to guide its predictions, while retaining an option to express uncertainty (NEI) when the available information was insufficient. However, defining a comprehensive intervention taxonomy resulted in a substantially longer prompt, increasing context usage and potentially reducing the context available for processing the clinical case. Two oncology experts independently assigned reference labels for all six evaluable cases, with disagreements resolved by consensus before analysis (Supplementary Section 6).

## 2.6 Expert Evaluation and Statistical Analysis

The commercial, proprietary, cloud-based comparator was ChatGPT-5.2, evaluated using the identical prompts provided in Supplementary Section 5, where applicable.

The outputs generated using the free-text recommendation strategy (as described in Section 2.5.1) were independently assessed by clinical experts against the Tier 1 treatment plans (ground truth). A structured evaluation instrument was developed to assess clinical correctness, adherence to NICE guidance, completeness, and patient safety. The questions presented to the assessors are provided in Supplementary Section 7.1. Based on the expert annotations, we calculated precision, recall, the Jaccard coefficient, overgeneration rate, miss rate, hallucination rate, and the proportion of additional interventions judged clinically appropriate (proportion appropriate). The Jaccard coefficient was included because precision and recall can each be maximised independently, whereas Jaccard provides a single measure of overall agreement between predicted and reference interventions. Overgeneration and miss rates are reported separately because Jaccard penalises false positives and false negatives equally, despite their potentially different clinical consequences. Proportion appropriate as a measure distinguishes clinically appropriate additional recommendations from true errors, recognising that LLMs may generate valid treatment options that are not explicitly represented in the MDT-derived reference plan.

For structured intervention prediction, assessors generated ground-truth labels for each Tier 1 case using the Intervention Taxonomy provided in Supplementary Section 6 and the corresponding MDT treatment plans. We used precision, recall, accuracy, F1 score, specificity, and Cohen's kappa to quantify model performance. The six Tier 1 cases were evaluated across 66 predefined interventions, resulting in 396 case-intervention prediction items per model. Because the structured output included a third class, Not Enough Information (NEI), performance was evaluated using three scoring schemes: NEI treated as No, NEI treated as Yes, and NEI treated as a separate third class. As the positive (Yes) class comprised approximately 10% of observations, three-class performance metrics were macro-averaged. Specificity was calculated using a one-versus-rest approach for the Yes class. Paired model comparisons used McNemar's test [55] for overall accuracy and paired bootstrap resampling (B=5,000) for class-specific performance metrics. Bootstrap resamples were generated by sampling the 396 paired case-intervention prediction items with replacement while preserving the pairing between models. The resulting dependence structure and its implications are described in Section 4.4.

## 2.7 Stakeholder Consultation

Two semi-structured group discussions were conducted with members of the HUTH breast MDT at Castle Hill Hospital: MDT coordinators ($n = 2$) and clinicians ($n = 12$; eight breast surgeons, two clinical oncologists, one radiologist, and one pathologist). Sessions were facilitated by WSJ, AD, and FH using a predefined topic guide covering MDT workflow, case preparation, documentation, and operational challenges. During the clinician session, the proposed system was demonstrated, and participants were invited to discuss its perceived benefits, risks, and implementation considerations. Notes were reviewed independently by WSJ, AD, and FH and synthesised to identify recurring themes relevant to workflow and system design, using an approach previously reported by this group [56, 57].

# 3. Results

## 3.1. ASR Robustness Under Acoustic Degradation

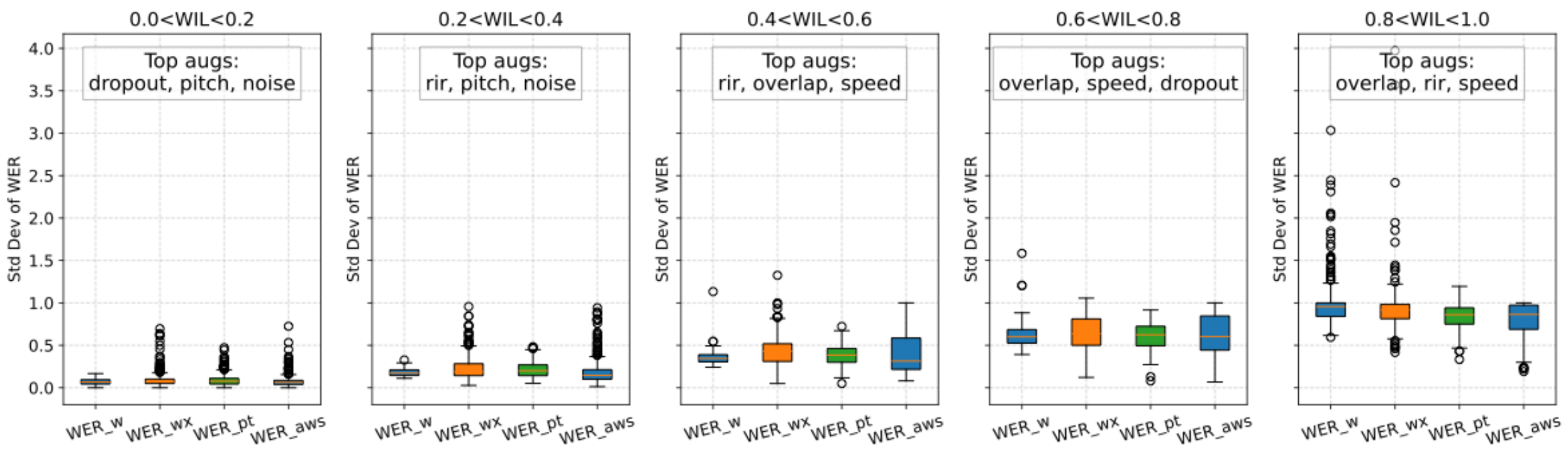

**Fig. 3** Distribution of word error rate across word information lost ranges and augmentation types, illustrating model robustness under increasing acoustic degradation

To examine ASR robustness under varying acoustic conditions, transcription performance of all the models was analysed across stratified levels of transcription degradation using WIL on Whisper large-v3 default configuration. Fig. 3 shows the distribution of raw WER values across five WIL ranges for all evaluated models. Across the 1,270 augmented recordings, WER remained consistently low for audio with WIL below 0.4, where background noise and pitch perturbation predominated. This suggests that contemporary ASR systems are generally robust to moderate environmental and spectral distortions. WER distributions widen substantially, with increased spread and emergence of higher-error instances between WIL values of 0.4 and 0.6, corresponding mainly to reverberation and partial speaker overlap. At WIL values above 0.6, interquartile ranges widened and extreme outliers became more frequent, indicating intermittent transcription failure rather than gradual performance decline. Recordings with WIL ≥ 0.8 were dominated by combinations of overlapping speech, speed perturbation, and reverberation. These findings suggest that ASR performance in MDT-like recordings is affected more by interacting temporal and multi-speaker distortions than by individual acoustic perturbations alone [58].

## 3.2. Model Selection for ASR

Table 1. Baseline automatic speech recognition performance on the augmented dataset (Tier 3), with clinical phrase recall. Amazon Transcribe Medical is a commercial benchmark, not a deployable component of the pipeline

| Metric | Mean WER | Mean WIL | Worst 3 WER (of 10 cases) | Clinical Phrase Recall |
|---|---|---|---|---|
| Whisper large-v3 | 0.39 | **0.44** | 3.03, 2.44, 2.05 | 0.56 |
| WhisperX | 0.41 | 0.47 | 3.97, 3.56, 1.94 | **0.60** |
| Parakeet | **0.37** | 0.46 | 1.19, 0.98, 0.96 | 0.52 |
| Amazon Transcribe Medical | 0.36 | 0.42 | 1.0, 1.0, 1.0 | 0.59 |

Baseline performance on Tier 3 dataset is summarised in for ASR

Table **1**. Amazon Transcribe Medical achieved the lowest WER and WIL, consistent with its optimisation for clinical speech. Among the open-source models, NVIDIA Parakeet achieved the lowest WER, whereas Whisper large-v3 achieved the lowest WIL. This ordering differed for clinically weighted evaluation, with Whisper large-v3 recovering more curated breast cancer diagnostic & treatment phrases than Parakeet (recall 0.56 vs 0.52). Although WhisperX achieved the highest phrase recall, it produced the highest WER and WIL and is based on the superseded Whisper large-v2 architecture; therefore, Whisper large-v3 was selected for subsequent optimisation.

Optimising Whisper large-v3 resulted in measurable improvements in transcription performance (optimised decoding parameters in Supplementary Table 6). A DNSMOS threshold of 3.5 on Tier 2 and Tier 3 dataset produced the lowest WER (0.3635) and WIL (0.4237), representing improvements of 7.3% and 4.5%, respectively, compared with the default configuration and reducing the performance gap with the commercial benchmark to 0.58% WER and 1.58% WIL. As optimisation was performed on the augmented dataset, these results represent in-sample performance.

**Table 2.** Effect of silence removal, hyperparameter optimisation, and DNSMOS-based adaptive decoding on transcription performance, repetition artefacts, and clinical terminology recall for the held-out authentic MDT recordings (Tier 1). Ground-truth repetition severity was 3.16 for Recording 1 (R1) and 0 for Recording 2 (R2). The complete configuration sweep is available in Supplementary Table 7.

| Configuration of Whisper large-v3 | Transcription Performance | Repetition Artefact Severity | Clinical Term Recall |
|---|---|---|---|

| | R1 | | R2 | | R1 | R2 | R1 | R2 |
|---|---|---|---|---|---|---|---|---|
| | WER | WIL | WER | WIL | | | | |
| Default | 0.62 | 0.72 | 0.27 | 0.34 | 9.11 | 0.80 | 0.31 | 0.63 |
| Silence removed | 0.59 | 0.72 | 0.26 | 0.36 | 15.83 | 0.88 | 0.22 | 0.52 |
| Tuned parameters | 0.58 | 0.74 | 0.24 | 0.38 | 1.82 | 0.32 | 0.25 | **0.66** |
| Silence removed + tuned | 0.53 | 0.69 | 0.22 | 0.33 | 3.10 | **0.00** | 0.25 | 0.64 |
| Tuned + threshold 2.5 | 0.70 | **0.52** | 0.33 | **0.26** | 4.09 | 0.67 | **0.32** | 0.61 |
| Silence removed + tuned + threshold 3.0 | 0.58 | 0.75 | **0.19** | 0.29 | 4.09 | 0.67 | **0.32** | 0.61 |
| Silence removed + tuned + threshold 3.5 | **0.49** | 0.66 | 0.21 | 0.33 | **0.62** | 0.15 | 0.28 | 0.58 |

Table 2 presents results on the held-out Tier 1 recordings. Silence removal combined with tuned decoding parameters reduced WER from 0.62 to 0.53 for Recording 1 and from 0.27 to 0.22 for Recording 2. Applying DNSMOS-based threshold selection at threshold of 3.5, further reduced WER to 0.49 for Recording 1 (20.7% reduction from the default configuration) and 0.21 for Recording 2 (24.4% reduction). The lowest WER for Recording 2 (0.19; 31.4% reduction) was achieved with a threshold of 3.0. Although the optimal threshold differed between recordings, values between 3.0 and 3.5 consistently produced the best performance.

Performance differed between the augmented (Tier 3) and authentic simulated (Tier 1) datasets. The optimised model achieved a WER on configuration *Silence removed + tuned + threshold 3.5* of 0.36 on the augmented audio compared with 0.49 and 0.21 on the two authentic MDT recordings, indicating greater variability in performance on representative MDT discussions. Owing to the limited number of authentic recordings, these observations should be interpreted cautiously.

Clinical term recall (Table 2) demonstrated that improvements in overall transcription accuracy did not consistently translate into improved preservation of clinically important terms. For Recording 1, the highest terminology recall (0.32) was 4.8% higher than the default configuration, whereas silence removal alone reduced recall from 0.31 to 0.22 despite improving WER. This divergence indicates that aggregate transcription metrics and clinical terminology preservation do not necessarily improve in parallel.

## 3.3. Case Extraction and Model Selection for Treatment Generation

Initial manual inspection of the outputs generated during case separation and treatment generation, without guideline context, showed that MedGemma produced more complete and consistently structured case summaries than Palmyra-Med. Palmyra-Med frequently summarised the discussions rather than extracting the clinical information and omitted content from longer MDT discussions. Although minor errors were observed in MedGemma-generated outputs, including terminology normalisation and occasional misclassification of individual sentences, the overall preservation of clinically relevant information was substantially better than with Palmyra-Med. Most inaccuracies appeared attributable to transcription noise rather than model hallucination, and the extracted outputs remained clinically interpretable.

Under the edge deployment constraints of limited unified memory, MedGemma was deployed as an 8-bit quantised 27B model, whereas Palmyra-Med required 3-bit quantisation to accommodate its 70B parameters, resulting in a greater reduction in model precision for Palmyra-Med. In addition, Palmyra-Med had a maximum output length of 4,096 tokens compared with 8,192 tokens for MedGemma. This shorter output limit increased the risk of truncation when processing longer MDT transcripts and limited the amount of clinical information that could be retained in the generated case description. Consequently, MedGemma was selected for subsequent recommendation generation.

## 3.4. Free-text Treatment Generation

**Table 3.** Performance comparison of LLM-based treatment prediction using retrieval-augmented generation (RAG). MR = MedGemma-RAG (local model); CR = ChatGPT-5.2 (cloud-based comparator). Prompt variants FT1 and FT2 correspond to the prompting strategies described in Supplementary Section 4.3.

| Model | Precision | Recall | Jaccard | Overgeneration | Miss rate | Hallucination | Proportion appropriate |
|---|---|---|---|---|---|---|---|
| MR - FT1 | 0.27 | 0.33 | 0.21 | 0.73 | 0.33 | **0.00** | **0.78** |
| CR - FT1 | 0.29 | **0.61** | 0.27 | 0.71 | **0.22** | **0.00** | 0.68 |
| MR - FT2 | **0.44** | 0.44 | **0.36** | **0.56** | 0.28 | 0.06 | 0.53 |
| CR - FT2 | 0.42 | 0.56 | 0.34 | 0.58 | 0.33 | 0.12 | 0.47 |

Table 3 summarises the performance of MedGemma-RAG (MR) and ChatGPT-5.2 (CR) against MDT-derived reference treatment plans under the two free-text prompt variants (FT1 and FT2). Across all model–prompt combinations, precision ranged from 0.27 to 0.44 and recall from 0.33 to 0.61. ChatGPT-5.2 achieved higher recall under both prompt variants, accompanied by higher overgeneration rates.

Under FT2, MR achieved higher precision (0.44 vs 0.42) and Jaccard coefficient (0.36 vs 0.34) than CR, while also producing fewer overgenerated interventions (0.56 vs 0.58). The proportion of additional interventions judged clinically appropriate was higher for MR (0.53 vs 0.47), and its hallucination rate was approximately half that of CR (0.06 vs 0.12). Under FT1, MR generated a higher proportion of clinically appropriate additional interventions than CR (0.78 vs 0.68). Neither model generated hallucinated interventions under this prompt variant. However, overgeneration remained common for both models, ranging from 0.56 to 0.73 across the two prompt variants. Between 47% and 78% of additional interventions generated by the models were judged clinically appropriate by expert reviewers. The FT2 reasoning-first prompt was associated with higher precision and lower overgeneration for MR, whereas FT1 was associated with a higher proportion of clinically appropriate additional interventions.

## 3.5. Intervention-based Structured Prediction

**Table 4.** Performance of intervention-based structured prediction across 396 paired predictions under three NEI scoring schemes. Scenario A treats NEI as No, Scenario B treats NEI as Yes, and the three-class analysis evaluates NEI as a separate class. Precision, recall, and F1 are reported for the Yes class in Scenarios A and B; the three-class analysis reports macro-averaged metrics with one-versus-rest specificity for the Yes class. Full paired-comparison statistics are provided in Supplementary Table 9.

| Scenario | Model | Accuracy | Precision | Recall | F1 | Specificity | Cohen's κ |
|---|---|---|---|---|---|---|---|
| A (NEI = No) | MR | 83.6% | 0.29 | 0.32 | 0.30 | 0.90 | 0.21 |
| A (NEI = No) | CR | 80.8% | 0.14 | 0.14 | 0.14 | 0.89 | 0.03 |
| B (NEI = Yes) | MR | 60.1% | 0.23 | 0.83 | 0.36 | 0.57 | 0.19 |
| B (NEI = Yes) | CR | 61.4% | 0.24 | 0.91 | 0.39 | 0.57 | 0.22 |
| Three-class | MR | 54.0% | 0.43 | 0.52 | 0.36 | 0.90 | 0.14 |
| Three-class | CR | 52.5% | 0.39 | 0.49 | 0.31 | 0.89 | 0.12 |

Table 4 compares MedGemma-RAG (MR) and ChatGPT-5.2 (CR) across 396 paired predictions (66 interventions across 6 cases) under the three scoring schemes described in Section 2.5.3. In Scenario A (NEI treated as 'No'), both models show high accuracy (MR 83.6%, CR 80.8%) and specificity due to class imbalance (~85% 'No'), but MR performs better on the clinically relevant 'Yes' class (F1: 0.30 vs 0.14; recall: 0.32 vs 0.14). Cohen's κ indicates slight-to-fair agreement for MR (0.21) and near-chance for CR (0.03). In Scenario B (NEI treated as 'Yes'), accuracy decreases (MR 60.1%, CR 61.4%) for both the models while recall increases (0.83 for MR and 0.91 for CR). In the 3-class setting, MR maintains a modest advantage (macro F1: 0.363 vs 0.314). MR, overall, performs slightly better, though results vary by scoring approach.

Given the relatively small sample size (n=396) and paired nature of the prediction items, paired statistical tests were used to assess differences in model performance. We used McNemar's test (MN) for accuracy and paired bootstrap (B = 5,000) (BS) for precision, recall, and F1, providing confidence intervals and effect size estimates without parametric assumptions. Formal comparison of MR and CR used McNemar's test for accuracy and a paired bootstrap (B = 5,000; n = 396) for class-specific metrics. See Supplementary Table 9 for detailed results of McNemar's test. Overall accuracy did not differ significantly under any scheme (Scenario A: +2.8 pp, p = 0.109; Scenario B: −1.3 pp, p = 0.630; 3-class: +1.5 pp, p = 0.561). However, class-stratified results showed clear differences. Under Scenario A (NEI treated as 'No'), MR significantly outperformed CR on the 'Yes' class: Precision +0.149 (95% CI [0.031, 0.275]; p = 0.016), Recall +0.182 (95% CI [0.041, 0.327]; p = 0.020), and F1 +0.165 (95% CI [0.038, 0.292]; p = 0.014). These differences persisted in the 3-class evaluation. On the 'No' class, MR showed a small advantage in Precision (+0.021; p = 0.014), with no differences in Recall or F1. Under Scenario B (NEI treated as 'Yes'), no significant differences were observed, as reclassification of NEI inflated CR's positive predictions. Both models performed poorly on the NEI class (F1 ≈ 0.08). MR's advantage is not reflected in accuracy but is consistent and specific to the clinically relevant 'Yes' class when abstention is not treated as a positive prediction.

## 3.6. Stakeholder Consultation

Four themes emerged across both stakeholder groups: fragmented digital infrastructure, administrative burden and workflow inefficiency, the potential value of AI-assisted documentation and data integration, and barriers to implementation relating to trust, governance, and system integration.

MDT coordinators described preparation as a multi-day process requiring manual collation of information from multiple non-interoperable hospital systems, followed by repeated transfer and cross-referencing into the Somerset Cancer Register [59] as the official MDT record. This duplication was viewed as time-consuming and cognitively demanding given the weekly caseload. Following demonstration of the proposed system, participants identified automated transcription, structured summarisation, and decision documentation as potential means of reducing administrative workload and improving documentation completeness, particularly when key personnel were unavailable. Concerns focused on the reliability of AI-generated outputs, integration with existing hospital systems, and the governance implications of recording MDT discussions.

Clinicians similarly identified fragmented patient information as a major barrier to efficient MDT decision-making, with relevant data distributed across electronic patient records, radiology, pathology, and MDT documentation systems. They considered automated generation of structured case summaries, prognostic scoring, guideline-informed treatment recommendations, and clinical trial identification to be the most valuable applications. Concerns centred on interoperability with existing systems, interpretation of unstructured clinical reports, and support for multiple clinically acceptable management options. Participants consistently emphasised that any AI system should undergo robust clinical validation, operate within clearly defined clinical boundaries, and integrate seamlessly into established MDT workflows.

# 4. Discussion

We developed and evaluated an end-to-end, edge-deployed AI pipeline that combines MDT speech capture with guideline-informed treatment recommendation while keeping all identifiable patient data within institutional infrastructure. A quantised 27B open-source model running on a single edge device outperformed a proprietary cloud-based model on the positive class of the structured prediction task, while the optimised open-source ASR pipeline achieved WIL within 1.58% of a commercial clinical benchmark. These findings demonstrate the technical feasibility of privacy-preserving, on-device AI for MDT decision support.

## 4.1. What ASR Performance Means in this Setting?

Aggregate transcription accuracy did not necessarily reflect clinically meaningful performance. Although NVIDIA Parakeet achieved a marginally lower WER than Whisper large-v3 (0.37 vs 0.39), Whisper recovered more clinically relevant terminology. In breast cancer MDTs, where receptor status, laterality, and disease stage determine treatment recommendations, preserving these entities is more important than overall lexical accuracy [60]. Configurations that improved WER sometimes reduced clinical terminology recall, indicating that WER alone is an insufficient optimisation target for clinical ASR [43]. Silence-aware preprocessing and DNSMOS-guided adaptive decoding substantially reduced long-form Whisper repetition artefacts without domain-specific retraining, highlighting the importance of preprocessing and inference strategy.

Unlike previous tumour board studies using curated case summaries, our pipeline processes spontaneous MDT discussions, introducing transcription uncertainty into downstream tasks. However, errors affecting negation, disease stage, and laterality are likely to have greater clinical impact than other transcription errors. Despite imperfect transcripts, the LLM frequently generated clinically coherent case descriptions, suggesting partial compensation for ASR errors during information extraction. Future work should address the methodological developments that can reduce error rate while improving the medical term recall. This may include optimising models on clinically significant transcription errors rather than relying solely on generic ASR metrics.

## 4.2. Guideline Grounding and Model Behaviour

Without RAG, both models generated incomplete treatment plans and unsupported guideline references, whereas retrieval of relevant NICE recommendations improved agreement with MDT decisions and reduced inappropriate recommendations, consistent with previous studies [25]. However, model behaviour varied substantially with both the underlying model and prompting strategy. ChatGPT-5.2 consistently achieved higher recall, capturing a larger proportion of MDT-relevant interventions, but also generated more additional interventions, indicating a broader suggestion-generation tendency and a less conservative prediction threshold. In contrast, MedGemma-RAG produced more targeted recommendations, achieving the best balance between precision and recall (Jaccard 0.3636) and higher agreement with MDT-derived plans under structured prompting.

The difference between FT1 and FT2 further demonstrates that prompt design can influence the balance between breadth and specificity [61]. Intervention-based prompting provided the models with a defined set of clinically relevant options, directing generation towards the interventions of interest and enabling more consistent quantitative evaluation. This suggests potential for adapting model behaviour to specific clinical tasks through task-specific prompt design rather than relying solely on the underlying model [62].

The edge deployment constraints also highlight an important trade-off. Although the NVIDIA Jetson AGX Orin provided sufficient unified memory to run the evaluated models, its memory limitations restricted the use of larger, higher-precision models and therefore the range of models that could be deployed locally. This trade-off is important for clinical deployment, where privacy requirements favour local inference but computational and financial resources remain constrained. Recent work suggests that smaller language models may offer a more practical and scalable alternative for healthcare applications because of their lower resource requirements, greater deployability and potential for task-specific optimisation [63]. Our findings therefore support a shift from simply selecting the largest available model towards optimising smaller models for well-defined, domain-specific clinical tasks. Such optimisation may provide a practical route to deploying capable, privacy-preserving AI within healthcare environments.

Nevertheless, guideline-based generation remains dependent on the completeness and relevance of the retrieved clinical context, and transcript-derived case descriptions cannot capture all information available to MDT members. Some interventions classified as overgeneration may therefore represent clinically appropriate options that were not explicitly verbalised during the discussion, consistent with the high proportion judged appropriate by expert reviewers. These limitations underscore the need to interpret model outputs alongside expert clinical judgement rather than as independent clinical recommendations.

## 4.3. Implementation Implications

Stakeholders identified documentation support and case triage as the most credible near-term applications of the proposed pipeline. Automated transcription, structured summarisation, and pre-meeting case preparation were viewed as valuable for reducing administrative burden, particularly the duplication associated with the Somerset Cancer Register. These findings complement the analytical validation by identifying clinical workflows in which the proposed pipeline could provide immediate benefit while supporting clinician-led decision-making.

Interoperability was identified as the principal implementation challenge. Because information is distributed across electronic patient records, radiology, pathology, and MDT documentation systems, AI will provide value only if integrated into existing clinical workflows. Although on-device processing reduces data transmission risks, governance issues relating to consent, data retention, and access control remain essential for deployment. Participants also emphasised the need for clearly defined clinical scope, robust validation, and seamless workflow integration.

For AIaMD, post-deployment monitoring should be considered during system design. The variability observed across authentic MDT recordings and the sensitivity of clinical terminology recall to decoding configuration indicate that performance may change with acoustic conditions, speaker composition, or guideline updates. Monitoring should therefore extend beyond aggregate ASR metrics to include preservation of clinically important entities and guideline currency.

## 4.4. Limitations

This study provides analytical validation and bench feasibility testing rather than clinical evaluation. The evaluation was based on six analysable cases from two authentic MDT recordings. Because multiple intervention predictions were generated from each case, the observations were clustered rather than independent, and the reported confidence intervals should be interpreted accordingly. All interventions were weighted equally, although missing chemotherapy is unlikely to have the same clinical consequence as omitting an axillary biopsy; future studies should incorporate consequence-weighted evaluation. Model calibration was not assessed, both models performed poorly on the NEI class, and interpretation of the reference labels should consider the reported inter-rater agreement (Section 3.5).

Furthermore, the RAG pipeline used a general-purpose embedding model, fixed top-*k* retrieval, and general-purpose ASR models without domain adaptation, all of which may limit performance. DNSMOS thresholding was applied at the recording level and may not reflect mixed acoustic conditions within recordings. Finally, the Tier 2 and Tier 3 datasets were generated from text-to-speech audio and did not reproduce the disfluencies and spontaneous turn-taking of authentic MDT discussions. To mitigate this, optimisation was validated on an independent held-out set of authentic recordings, and synthetic and authentic performance are reported separately.

### 4.5. Future Work

Future work will focus on improving technical performance through domain-specific ASR models and Graph RAG-based treatment recommendation, supported by larger datasets for model adaptation and evaluation. Integration with electronic health records will be essential for the documentation workflows identified by stakeholders. Before recommendation-facing deployment, further work should address consequence-weighted evaluation, calibration, uncertainty estimation, explainability, post-deployment monitoring, and safety assurance. Guided by the DECIDE-AI framework, prospective evaluation in live MDT meetings will assess clinical performance, workflow integration, and user acceptance.

## 5. Conclusions

We developed and evaluated an edge-deployed AI pipeline using open-source ASR and LLM models that integrates ASR, locally hosted LLMs, and guideline-informed retrieval to support breast cancer MDT documentation and treatment recommendation without transmitting identifiable patient data beyond institutional infrastructure. The optimised ASR pipeline achieved performance within 1.58% (WIL) of a commercial clinical benchmark on augmented audio, while MedGemma-RAG identified 2.3 times more MDT-concordant interventions than a proprietary cloud-based comparator with consistently low hallucination rates. Performance was more variable on authentic MDT conversations than on synthetic audio, highlighting the importance of evaluating conversational clinical data. Differences between prompting strategies also demonstrate that task-specific prompt design can influence the balance between treatment-plan coverage and concordant to MDT decisions. These findings support the feasibility of using smaller, locally deployable models for domain-specific clinical tasks, while highlighting the need to optimise model and prompt design for constrained edge environments. Stakeholder consultation indicated that documentation support and case preparation represent the most credible near-term applications, with workflow integration and governance emerging as greater barriers to implementation than model performance. Prospective clinical evaluation, consequence-weighted assessment, and post-deployment monitoring are required before routine clinical use.

## Statements and Declarations

**Acknowledgements:** The authors thank the members of the Hull University Teaching Hospitals NHS Foundation Trust breast cancer multidisciplinary team who contributed simulated case discussions and participated in the stakeholder consultation sessions.

**Competing interests:** The authors have no financial or non-financial competing interests to declare that are relevant to the content of this article.

**Funding:** This study was supported by a grant from the Humber and North Yorkshire Cancer Alliance, Cancer Research and Innovation Award Funding 25-26.

**Ethics approval:** Ethical approval was not required for this study. The study involved simulated MDT discussions conducted by members of the HUTH breast cancer MDT using fully de-identified clinical scenarios. The exercise was undertaken specifically for the purposes of this study and was performative in nature, with clinicians discussing the presented scenarios and formulating corresponding management decisions. The stakeholder consultation comprised focus group discussions on the breast cancer MDT pathway and the proposed technology. Participants provided informed consent to participate and to audio recording of these discussions, and no sensitive personal or patient-identifiable information was collected.

**Consent to participate:** Informed consent was obtained from all MDT members who participated in the recorded simulated-case discussions and stakeholder consultation sessions.

**Consent to publish:** Not applicable, as no identifiable participant or patient information is included in this manuscript.

**Data availability:** The datasets generated and/or analysed during the current study are available from the corresponding author on reasonable request, subject to institutional governance and data protection requirements.

**Code availability:** The code used to develop and evaluate the pipeline is available from the corresponding author on reasonable request.

**Author contributions:** Conceptualisation: WSJ, FH, KG, AD; Methodology: AD, WSJ, LBS, FH; Software: AD, LBS; Analysis and Evaluation: AD, FH, IM, WSJ; Data curation: AD, WSJ, FH; Manuscript- original draft preparation: WSJ, AD; Manuscript - review and editing: all authors.

**Use of AI tools:** All AI and LLM tools used for analytic or generative tasks are disclosed in Methods and Supplementary Material

**Clinical Trial Number:** not applicable.

# Supplementary Information

**Article Title:** Development and Feasibility Evaluation of an Edge AI as Medical Device System for Breast Cancer Multidisciplinary Team Meetings

**Authors:** Aarzoo Dhiman[1] (A.Dhiman@hull.ac.uk), Farzana Haque[2] (Farzana.Haque4@nhs.net), Iqtedar Muazzam[2] (iqtedar.muazzam@nhs.net), Kartikae Grover[2] (kartikaegrover@nhs.net), Lydia Brian Smith[3] (L.Bryan-Smith-2014@hull.ac.uk), William Stephen Jones[1] (Will.Jones@hull.ac.uk)

**Corresponding author:** Aarzoo Dhiman, Centre of Excellence for Data Science, Artificial Intelligence and Modelling (DAIM), Faculty of Science and Engineering, University of Hull, Hull, United Kingdom. Email: a.dhiman@hull.ac.uk

## Supplementary Section 1: Rapid Scoping Review: Search Strategy and Evidence Table

### 1.1 Search Strategy

Database: PubMed/MEDLINE.

Copy-paste executable search string:

("breast cancer" AND ("multidisciplinary team" OR MDT OR "tumor board" OR "tumour board")) AND ("artificial intelligence" OR "machine learning" OR "clinical decision support" OR "large language model" OR LLM OR NLP OR "retrieval augmented generation")

A parallel search was performed for automatic speech recognition studies in breast cancer MDT settings.

### 1.2 Evidence Table: LLM, RAG and AI Decision Support in Breast Cancer MDT Settings

**Supplementary Table 1.** Characteristics and reported performance of published AI systems for breast cancer MDT decision support, including system role, study design, evaluation approach, and key outcomes.

| Author(s), year | AI system | Role in MDT | Design and evaluation | Outcome |
|---|---|---|---|---|
| Sorin et al., 2023 [1] | ChatGPT-3.5 using clinical case information | Treatment recommendation and case summarisation for breast tumour | Proof-of-concept comparison with tumour board decisions (10 breast cancer cases) | 70% agreement with tumour board decisions |

| Author(s), year | AI system | Role in MDT | Design and evaluation | Outcome |
|---|---|---|---|---|
| | | board | | |
| Griewing et al., 2023 [2] | ChatGPT-3.5 generating treatment recommendations | Compared with breast cancer tumour board decisions | Observational comparison using diverse breast cancer patient profiles | ~50% overall concordance; 58.8% for invasive cases |
| Griewing et al., 2024 [3] | GPT-4, GPT-3.5, Llama2, Bard | Treatment recommendation support for MDT decisions | Comparison with MDT recommendations, 20 complex breast cancer cases | GPT-4 highest concordance (70.6%); other models lower |
| Xu et al., 2024 [4] | CSCO AI clinical decision support system | Treatment recommendation support for MDT | Retrospective study of 537 breast cancer patients | 92.4% concordance with MDT recommendations |
| Liao et al., 2025 [5] | ChatGPT-4.0 for breast cancer case recommendations | Compared with expert tumour board recommendations | 362 breast cancer cases evaluated by expert tumour board and ChatGPT | 46% concordance with experts; 39% reproducibility across responses |
| Dogan et al., 2025 [6] | ChatGPT-4.0 using clinical case summaries | Treatment recommendation support | Prospective comparison with MDT decisions (100 cancer cases) | 76.4% agreement with MDT decisions |
| Ah-Thiane et al., 2025 [7] | Claude 3 Opus, GPT-4 Turbo, LLaMA3-70B | Treatment recommendation support for MDT decisions | Retrospective comparison with expert MDT decisions (112 early breast cancer cases) | 86.6% (Claude 3 Opus), 85.7% (GPT-4 Turbo), 75% (LLaMA3-70B) |
| Büyükceran et al., 2025 [8] | GPT-4o generating treatment plans from structured clinical data | Decision support for breast cancer MDT | Retrospective comparison with MDT decisions (33 patients) | 93.9% full concordance with MDT decisions |
| Umihanic et al., 2025 [9] | ChatGPT-4.0 generating treatment recommendations from patient data | Decision support for breast cancer MDT | Retrospective comparison with MDT decisions (91 patients) | High agreement (mean score 3.31/4); better in standard cases |
| Schmutz et al., 2025 [10] | ChatGPT-4.0 generating therapy recommendations from molecular | Decision support for molecular tumour board | Retrospective comparison with expert MTB decisions (20 cancer cases) | More treatment suggestions and faster recommendations; |

| Author(s), year | AI system | Role in MDT | Design and evaluation | Outcome |
|---|---|---|---|---|
| | case data | | | moderate consistency (κ ≈ 0.51) |
| Zhou et al., 2021 [11] | IBM Watson for Oncology | Treatment recommendation support for MDT | Meta-analysis of 9 studies comparing WFO and MDT decisions (2,463 patients) | 81.5% overall concordance with MDT decisions |
| Moser & Narayan, 2020 [12] | AI-driven predictive care-coordination toolkit | Supports MDT care planning and patient management | Conceptual/implementation discussion | May improve care coordination, risk prediction and patient management; requires human oversight |

## Supplementary Section 2: Datasets, Models and Full Configuration Results

### 2.1 Tier 2 Synthetic Voice Configuration (Amazon Polly)

**Supplementary Table 2** Amazon Polly voices used to generate synthetic MDT discussions, showing the assigned MDT role, voice profile, language variety, gender, and synthesis type.

| MDT role | Polly voice | Accent / language variety | Gender | Neural/Standard |
|---|---|---|---|---|
| Radiologist | Arthur | British English | Male | Neural |
| Pathologist | Raveena | Indian English | Female | Standard |
| Surgeon | Emma | British English | Female | Neural |
| Oncologist | Aditi | Indian bilingual English | Female | Standard |
| Breast care nurse | Kajal | Indian bilingual English | Female | Neural |
| MDT coordinator | Brian | British English | Male | Neural |

### 2.2 Tier 2 Dataset Composition

**Supplementary Table 3** Characteristics of the ten synthetic MDT discussion scenarios, including discussion duration and multidisciplinary speaker roles represented in each case.

| Case ID | Duration (s) | Speaker roles represented |
|---|---|---|
| 1 | 29 | Radiologist, pathologist, surgeon, MDT coordinator |

| Case ID | Duration (s) | Speaker roles represented |
|---|---|---|
| 2 | 38 | Radiologist, pathologist, surgeon, oncologist, breast care nurse, MDT coordinator |
| 3 | 60 | Radiologist, pathologist, surgeon, oncologist, MDT chair (consultant surgeon), MDT coordinator |
| 4 | 29 | Oncologist, radiologist, pathologist, palliative care specialist, MDT coordinator |
| 5 | 30 | Radiologist, pathologist, surgeon, MDT coordinator, chair |
| 6 | 44 | Radiologist, pathologist, oncologist, surgeon, breast care nurse, MDT coordinator, chair |
| 7 | 38 | Radiologist, pathologist, surgeon, oncologist, chair, MDT coordinator |
| 8 | 29 | Radiologist, pathologist, surgeon, breast care nurse, MDT coordinator, chair |
| 9 | 23 | Surgeon, pathologist, oncologist, MDT coordinator, chair |
| 10 | 32 | Radiologist, pathologist, oncologist, palliative care specialist, breast care nurse, MDT coordinator, chair |

## 2.3 Acoustic Augmentation Parameters

Each of the ten Tier 2 synthetic recordings was augmented across all combinations of the techniques below, producing 127 variants per recording and 1,270 augmented files in total. Severely degraded configurations were retained deliberately to permit stress testing at the limits of intelligibility.

**Supplementary Table 4** Acoustic data augmentation techniques applied to synthetic MDT recordings, including implementation methods and parameter ranges used to generate the augmented evaluation dataset.

| Technique | Implementation | Parameter values |
|---|---|---|
| Background noise | Mixing with environmental recordings (keyboard typing, office ambience, paper handling, small-room conversation, crowded meeting-room speech) scaled to target SNR | SNR 0, 5, 10, 15 dB |
| Room impulse response | Convolution with RIR filters to replicate meeting-room reverberation | audio files with RIR used |
| Speech overlap | Insertion of randomly selected segments from secondary audio files into the primary signal | Overlap ratio 0.1, 0.15, 0.3; segment length 1–20 s; 2–10 segments per recording; mixing volume ratio 0.1, 0.4, 0.8 |
| Speed perturbation | Resampling to produce faster/slower versions | Speed factors 0.95, 1.3, 1.5, 2.0 |
| Pitch shifting | Pitch shift without duration change | −2, 0, +2 semitones |
| Audio dropout | Chunk-based zeroing of waveform segments to | Dropout probability 0.05, 0.1 |

| Technique | Implementation | Parameter values |
|---|---|---|
| | simulate recording interruption | |

All augmented audio was produced in the same format as the source synthetic audio (16 kHz, 16-bit, mono).

## 2.4 ASR Models Evaluated

**Supplementary Table 5** Automatic speech recognition (ASR) systems evaluated in this study, summarising model characteristics, intended capabilities, and role within the evaluation pipeline.

| Model | Description |
|---|---|
| Whisper large-v3 | Open-source encoder–decoder ASR model (~1.55B parameters) supporting multilingual transcription and translation, using 128 Mel-frequency filter banks, with reported robustness to background noise, overlapping speech and domain-specific terminology |
| NVIDIA Parakeet | Family of GPU-optimised ASR models (0.6–1.1B parameters) designed for high-throughput inference, with competitive WER, faster inference than Whisper, and native punctuation and timestamp generation |
| WhisperX | Extension of Whisper adding phoneme-level alignment and speaker diarization, improving timestamp precision and enabling multi-speaker attribution, with batched inference and alignment correction via secondary models. Built on Whisper large-v2 |
| Amazon Transcribe Medical | HIPAA-eligible cloud ASR service for clinical documentation, with strong medical vocabulary support. Requires cloud processing and incurs ongoing operational cost, and was therefore used as a benchmark only |

## 2.5 Whisper Large-v3 Decoding Parameters Optimised

Optimised via Optuna (Tree-structured Parzen Estimator, 100 trials, parallel search, objective: minimise WER on Tier 3 augmented audio): temperature, beam_size, best_of, compression_ratio_threshold, logprob_threshold, length_penalty, hallucination_silence_threshold.

**Supplementary Table 6** Optimised Whisper large-v3 decoding parameters identified through Bayesian hyperparameter optimisation and used for adaptive transcription.

| Parameter name | Optimised Value |
|---|---|
| temperature | 0.7596350092541306 |
| beam_size | 2 |
| best_of | 2 |
| compression_ratio_threshold | 1.5323007177495684 |
| logprob_threshold | -1.0903724240294121 |
| length_penalty | 0.7594738864766823 |
| hallucination_silence_threshold | 0.22902235239020646 |

Silence detection: FFmpeg silencedetect filter, −23 dB threshold, segments longer than 5s removed with 2s of surrounding silence retained before concatenation.

### 2.6 Full DNSMOS Threshold Sweep on Authentic Recordings (Tier 1)

**Supplementary Table 7** Word error rate (WER) and word information lost (WIL) of Whisper large-v3 across default and optimised transcription configurations on the two held-out authentic MDT recordings (Recording 1: R1 and Recording 2: R2).

| Model configuration | R1 WIL | R1 WER | R2 WIL | R2 WER |
|---|---|---|---|---|
| Whisper default | 0.7252 | 0.6181 | 0.3412 | 0.2738 |
| Default + silence removed | 0.7184 | 0.5938 | 0.3567 | 0.2649 |
| Tuned parameters | 0.7451 | 0.5761 | 0.3813 | 0.2445 |
| Silence removed + tuned | 0.6941 | 0.5342 | 0.3265 | 0.2210 |
| Tuned + noise threshold 2.5 | 0.7005 | 0.5221 | 0.3293 | 0.2624 |
| Tuned + noise threshold 3.0 | 0.7473 | 0.5651 | 0.3483 | 0.2337 |
| Tuned + noise threshold 3.5 | 0.6971 | 0.5198 | 0.3467 | 0.2223 |
| Silence removed + tuned + threshold 2.5 | 0.6900 | 0.5088 | 0.3457 | 0.2726 |
| Silence removed + tuned + threshold 3.0 | 0.7474 | 0.5827 | 0.2953 | 0.1878 |
| Silence removed + tuned + threshold 3.5 | 0.6571 | 0.4901 | 0.3281 | 0.2070 |

## Supplementary Section 3: ASR Evaluations

### 3.1 Evaluation Metric

$$\mathrm{WER} = \frac{\mathrm{S} + \mathrm{D} + \mathrm{I}}{\mathrm{N}}$$

Where: S = number of substitution, D = number of deletions, I = number of insertions, and N = total number of words in the reference (ground truth).

$$\mathrm{WIL} = 1 - \frac{\mathrm{H}^2}{\mathrm{N} \cdot \mathrm{M}}$$

Where: H = number of correctly recognised words (hits), N = total number of words in the reference, M = total number of words in the hypothesis.

### 3.2 Curated Clinical Phrase List for ASR Terminology Evaluation

Phrases were extracted from the reference transcripts of each case and matched against ASR output using exact and fuzzy (Levenshtein similarity ≥ 80%, FuzzyWuzzy) comparison of 1–6-grams. Recall was computed as TP / (TP + FN). Precision was not computed because candidate terms were restricted to this predefined vocabulary.

**Case 1:** 56; routine screening; 8 mm; lesion; right upper outer quadrant; invasive ductal carcinoma; BI-RADS 5; core biopsy; grade 1; ER positive; HER2 negative; straightforward lumpectomy; nodal involvement; wide local excision; sentinel node biopsy

**Case 2:** 42; palpable lump; MRI; multifocal disease; 6 cm; axillary nodes; biopsy; triple negative invasive carcinoma; grade 3; nodal core also positive; too extensive; breast conserving surgery; neoadjuvant chemotherapy; anxious; downstage; reassess surgical options

**Case 3:** 68; 61; male breast cancer; imaging; 2 cm lesion; no nodes; ER positive; HER2 negative; simple mastectomy; endocrine therapy; surgery if fitness is confirmed; survival benefit; anaesthetic review; surgical referral

**Case 4:** 72; surgery; chemotherapy; liver metastases; follow-up scans; bilateral liver lesions; not resectable; HER2 positive; HER2 targeted therapy; prognosis is guarded; plan supportive care; referral to palliative team; close monitoring

**Case 5:** 56; routine screening; 8 mm lesion; right breast; upper outer quadrant; malignant; BI-RADS 5; core biopsy; invasive ductal carcinoma; grade 1; ER positive; HER2 negative; wide local excision; sentinel node biopsy; lumpectomy

**Case 6:** 42; palpable lump; symptomatic; MRI; multifocal disease; 6 cm span; abnormal axillary nodes; triple negative invasive carcinoma; grade 3; nodal core also positive; neoadjuvant chemotherapy; breast conserving surgery not feasible; anxious; reassess surgery after response; psychological support; breast care team

**Case 7:** 68; male breast cancer; 2 cm lesion; no nodes; ER positive; HER2 negative; simple mastectomy; endocrine therapy; COPD; cardiac history; surgery risk is high; offers better local control; refer to anaesthetics; surgical referral; endocrine therapy if unfit

**Case 8:** 33; family history; genetic screen positive; MRI; 6 mm lesion; right breast; biopsy; high grade DCIS; BRCA1 mutation; bilateral mastectomy; risk reducing surgery; refer to genetics; plan for bilateral mastectomy; reconstruction discussion

**Case 9:** 61; lumpectomy; tumour 1.2 cm; grade 2; ER positive; HER2 negative; margins clear; no nodal involvement; adjuvant endocrine therapy; no chemotherapy needed; endocrine therapy; routine follow-up

**Case 10:** 72; post surgery; chemotherapy; multiple liver metastases; HER2 positive; HER2 targeted therapy; prognosis is poor; needs early palliative involvement; symptom control; fatigue; mobility issues; palliative care referral; supportive measures

**NOTE.** Spellings have been corrected relative to the original working list (“gaurded” → “guarded”, “reasses” → “reassess”, “BIRADS” → “BI-RADS”, “HER 2” → “HER2”). If the matching code used the original strings verbatim, either re-run matching with the corrected list or state explicitly that matching used the working spellings, since fuzzy matching at 80% similarity is sensitive to these variants.

## Supplementary Section 4: LLM Decoding Hyperparameters

**Supplementary Table 8** Inference parameters used for MedGemma 27B and Palmyra-Med 70B during case extraction and treatment separation.

| Model | Parameters (case extraction and treatment separation) |
| --- | --- |
| MedGemma 27B | temperature 0.05; top_p 0.7; min_p 0.05; repeat_penalty 1.2; frequency_penalty 0.3; typical_p 0.7; top_k 20; mirostat_mode 0; tfs_z 1.0; max_tokens 4000 |
| Palmyra-Med 70B | temperature 0.02; top_p 0.6; min_p 0.15; repeat_penalty 1.15; frequency_penalty 0.1; typical_p 0.7; top_k 15; mirostat_mode 0; tfs_z 1.0 |

For guideline-grounded treatment generation, MedGemma was run with fixed parameters: temperature 0.1; top_p 0.9; repeat_penalty 1.1; top_k 40; max_tokens 1000.

## Supplementary Section 5: Prompting Strategy Used

### 5.1 Prompt: Case Description and Treatment Separation

```
You are a clinical MDT transcription assistant specialized in breast oncology.
Your tasks:

1. Consider the given text as the MDT meeting discussion transcript of a single case.

2. For the given case, you may clean and normalise the language by correcting spelling, grammar and punctuations without including/excluding any information. with your breast cancer expertise, you have to clean the medical terminologies wherever possible, by correcting ASR phonetic drift into medically valid terms.

3. For the case, you have to extract the case description, you MUST NOT include or modify any information not found in the given text. You HAVE TO only use the information provided in the text with corrections.

4. The case description may include patient’s features/conditions like, age, gender, demographics, diagnosis, pathology, imaging, nodal status, biomarkers etc.

5. For each case, you also have to extract the Treatment Decisions made during the MDT meeting, using this text. you MUST NOT include or modify any information not found in the given text. You HAVE TO only use the information provided in the text.

6. The treatment decisions may include agreed MDT plan, investigations, referrals, follow-up actions etc.

7. For both case description and treatment plan, you have to use the EXACT VERBATIM from the provided text, without any structural modifications.

8. IF the details are not clear, mark it as “UNCLEAR in transcript”.

Output format (STRICT):
```

```
For each case, use the following structure:

CASE X:

Corrections made:

Case Description:

Treatment Decisions:

===============================

EXAMPLE:

!---------------Example not provided for data privacy-------------!

"""
```

A one-shot worked example was appended to this prompt, comprising a sample noisy transcript excerpt and the expected corrected output.

### 5.2 Prompt: Treatment Prediction without Guideline Context

```
SYSTEM ROLE:

You are a clinical decision-support assistant for multidisciplinary team (MDT) meetings.

You do NOT provide medical advice.

You summarize cases and map them to high-level NICE guideline pathways.

TASK:

From the transcript below:

1. Understand the case description carefully, ONLY consider information that is present.

2. Propose a TREATMENT PLAN that is CONSISTENT WITH NICE BREAST CANCER GUIDELINES

RULES:

- Use ONLY information explicitly stated in the transcript

- Do NOT assume staging, performance status, or comorbidities

- Do NOT invent biomarkers, imaging results, or pathology

- If required information is missing, write: INSUFFICIENT INFORMATION

- Treatment plans must be guideline-aligned, not personalized prescriptions

- Use conditional language: "may be considered", "is typically recommended", "depending on eligibility"

OUTPUT FORMAT:

  "Case Description":
```

```
  "Proposed Treatment Plan":

  "guideline basis":

  "confidence": "low | medium | high"

IMPORTANT:

- If multiple patients are present, output an ARRAY (one per case)

- If no clear treatment decision can be derived, populate treatment_plan with
"INSUFFICIENT INFORMATION"
```

---

```
==============================================

EXAMPLE:

!---------------Example not provided for data privacy-------------!

"""
```

## 5.3 Prompt: Treatment Prediction with Retrieved Guideline Context (RAG)

### 5.3.1 Free-text Treatment Prediction 1 (FT1)

```
f"""

SYSTEM ROLE:

You are a clinical decision-support assistant for multidisciplinary team (MDT)
meetings.

You do NOT provide medical advice.

You summarize cases and map them to high-level NICE guideline pathways.

TASK:

From the transcript below:

1. Understand the case description carefully, ONLY consider information that is
present.

2. Propose a TREATMENT PLAN that is CONSISTENT WITH NICE BREAST CANCER GUIDELINES

RULES:

- Use ONLY information explicitly stated in the transcript

- Do NOT assume staging, performance status, or comorbidities

- Do NOT invent biomarkers, imaging results, or pathology

- If required information is missing, write: INSUFFICIENT INFORMATION

- Treatment plans must be guideline-aligned, not personalized prescriptions

- Use conditional language: "may be considered", "is typically recommended",
"depending on eligibility"

OUTPUT FORMAT:
```

```
"Case Description":

"Proposed Treatment Plan":

"guideline basis":

confidence": "low | medium | high"
```

IMPORTANT:

- If multiple patients are present, output an ARRAY (one per case)

- If no clear treatment decision can be derived, populate treatment_plan with "INSUFFICIENT INFORMATION"

==============================================

EXAMPLE:

!---------------Example not provided for data privacy-------------!

"""

**5.3.2 Free-text Treatment Prediction 2 (FT2)**

SYSTEM ROLE:

You are a clinical decision-support assistant for multidisciplinary team (MDT) meetings.

You do NOT provide medical advice.

Extract the MDT's explicit clinical reasoning and decision nodes, and align each to relevant NICE guidance where applicable.

TASK:

From the transcript below:

1. Extract the MDT's explicit clinical reasoning and key decision drivers.
2. Reconstruct the proposed treatment strategy exactly as discussed (including sequencing, conditional plans, and referrals).
3. Then verify that the strategy is consistent with relevant NICE guidance.

MDT REASONING REQUIREMENTS:

In your Proposed Treatment Plan:

- Explicitly identify timing decisions (e.g., neoadjuvant vs adjuvant)
- Explicitly identify biomarker-dependent branching (e.g., PD-L1, BRCA)
- Explicitly identify surgical morbidity trade-offs if discussed
- Explicitly identify when genetic results influence surgical planning
- Explicitly identify MDT referrals (e.g., bone MDT)
- Explicitly identify when staging is NOT required
- Preserve uncertainty or conditional logic expressed by the MDT
- Do NOT collapse strategic reasoning into generic pathway summaries.

```
RULES:

- Use ONLY information explicitly stated in the transcript

- If the MDT explicitly rejects a standard step (e.g., "no staging required"), preserve that decision and do not reintroduce it.

- Do NOT invent biomarkers, imaging results, or pathology

- If required information is missing, write: INSUFFICIENT INFORMATION

- Treatment plans must be guideline-aligned, not personalized prescriptions

- Use conditional language: "may be considered", "is typically recommended", "depending on eligibility"

- If the transcript contains biomarker testing or pending results, structure the treatment plan with explicit IF/THEN branches.

- Do NOT default to surgery-first or adjuvant-first pathways unless explicitly supported by the transcript.

OUTPUT FORMAT:

  "Case Description":

  "Proposed Treatment Plan":

 "Key Decision Drivers":

- (bullet points listing what determined the plan)

  "guideline basis":

  "confidence": "low | medium | high"


IMPORTANT:

- If multiple patients are present, output an ARRAY (one per case)

- If no clear treatment decision can be derived, populate treatment_plan with "INSUFFICIENT INFORMATION"
```

---

```
EXAMPLE:

!---------------Example not provided for data privacy-------------!

"""
```

Retrieved context: the top five guideline chunks returned by FAISS similarity search over the embedded NICE corpus were inserted into the prompt. Generation used fixed decoding parameters (temperature 0.1, top_p 0.9, repeat_penalty 1.1, top_k 40, max_tokens 1000).

### 5.4 Prompt: Intervention-based Structured Prediction

```
SYSTEM ROLE:
```

You are a clinical decision-support assistant for multidisciplinary team (MDT) meetings.

Extract the MDT's explicit clinical reasoning and decision nodes, and align each to relevant NICE guidance where applicable.

TASK:

From the transcript below:

1. Understand the case description carefully and the context from the NICE guidelines, ONLY consider information that is present.

2. You will be given a list of possible treatment options that may or may not be applicable to the case description. Your task is to label the treatment option as (i) YES (ii) NO, and (iii) NEI (Not Enough Information) with a brief outline for the reason of the label made.

3. Then verify that the strategy is consistent with relevant NICE guidance.

Below is the list of treatment options that you have to label. Use the full list, do not add or remove any option from the list while doing the labelling.

Treatment options (label each as YES/NO/NEI):

1. Surgical Procedures:

1.1. Diagnostic & Image-Guided Procedures

1.1.1. Biopsies (core, vacuum, excisional, punch)

1.1.2. Localisation techniques (wire, seed, magnetic, ultrasound-guided)

1.2. Breast-Conserving Surgery (BCS)

1.2.1. Lumpectomy / Wide Local Excision

1.2.2. Margin re-excision

1.2.3. Oncoplastic Breast-Conserving Surgery

1.3. Mastectomy

1.3.1. Simple, skin-sparing, nipple-sparing

1.3.2. Modified radical

1.3.3. Skin-reducing mastectomy

1.4. Risk-Reducing & Prophylactic Surgery

1.4.1. Bilateral risk-reducing mastectomy

1.4.2. Contralateral prophylactic mastectomy

1.5. Axillary Staging, treatment

1.5.1. Sentinel lymph node biopsy (SLNB)

1.5.2. Axillary lymph node dissection (ALND)

1.5.3. Targeted axillary procedures

1.6. Breast Reconstruction (Oncoplastic & Post-Mastectomy)

1.6.1. Implant-based reconstruction

1.6.2. Autologous flaps (DIEP, LD)

1.6.3. Immediate & delayed reconstruction

1.7. Nipple/Areola Reconstruction

1.7.1. Nipple reconstruction (local flaps, grafts)

1.7.2. Areola tattooing

1.8. Recurrent & Palliative Breast Surgery

1.8.1. Chest wall resection

1.8.2. Surgery for recurrence

1.8.3. Palliative procedures

2. Imaging

2.1. Mammography

2.2. Contrast-Enhanced Mammography (CEM)

2.3. Breast Ultrasound (US)

2.4. Breast MRI

2.5. FDG Positron Emission Tomography-CT (PET-CT)

2.6. Sodium Fluoride PET-CT

2.7. CT Scan of Chest/Abdomen/Pelvis

2.8. Bone Scan (Scintigraphy)

3. Radiotherapy

3.1. Adjuvant Whole Breast Irradiation/ Chest Wall Irradiation (post-mastectomy radiotherapy, PMRT)

3.2. Tumour Bed Boost

3.3. Partial Breast Irradiation (PBI / APBI)

3.4. Regional Nodal Irradiation (RNI)

3.4.1. Axillary irradiation

3.4.2. Supraclavicular nodal irradiation

3.4.3. Internal mammary node (IMN) irradiation

3.5. Stereotactic & Special Techniques- Stereotactic body radiotherapy (SBRT) for oligometastases

3.6. Palliative Radiotherapy

3.6.1. Palliative breast/chest wall irradiation

3.6.2. Radiotherapy for bone metastasis

3.6.3. Radiotherapy to brain metastasis (WBRT / SRS)

3.7. No need for radiotherapy

4. Systemic treatment

4.1. Chemotherapy- Neoadjuvant/adjuvant/palliative chemotherapy

4.2. Endocrine (Hormonal) Therapy

4.2.1. Anti-estrogen therapy- Tamoxifen, Aromatase inhibitors

4.2.2. Ovarian suppression/ablation

4.3. Targeted Therapy (Biologic Therapy)

4.3.1. HER2-targeted therapy

4.3.2. CDK4/6 inhibitors

4.3.3. PI3K / mTOR inhibitors

4.3.4. Antibody-drug conjugates (ADCs)

4.3.5. PARP inhibitors (For BRCA-positive patients)

4.3.6. NTRK targeted (NTRK fusion positive patients)

4.4. Immunotherapy

4.5. Bone-Targeted Therapy- Bisphosphonates, RANKL inhibitors (denosumab)

4.6. Electrochemotherapy

4.7. Supportive treatments

5. Useful Tools

5.1. Prognostic & Treatment Benefit Calculators

5.1.1. PREDICT tool

5.1.2. Nottingham Prognostic Index (NPI)

5.2. Biomarkers

5.2.1. ER / PR status

5.2.2. HER2 status

5.2.3. Ki-67 proliferation index

5.2.4. PD-L1 expression

5.3. Genomic / Molecular Assays- Oncotype Dx

5.4. Common terms used- TNM staging, Tumour grade (Nottingham grading system), Nodal status

6. Referrals

6.1. Referral to other MDT

6.2. Refer to Oncology Health

6.3. Refer for clinical trials

7. Appointments

7.1. Clinical Oncology appointment

7.2. Medical Oncology appointment

LABELING RULES:

Answer YES if the option is explicitly: should be (done/performed OR recommended/planned as next management, OR requested/accepted as the intended plan) AND aligned with (the NICE guidelines, OR per your existing knowledge).

Answer NO if the option is:

should NOT be (done/performed OR recommended/planned as next management, OR requested/accepted as the intended plan) OR NOT aligned with (the NICE guidelines, OR per your existing knowledge). OR should be explicitly rejected/declined/ruled out.

Answer Not Enough Information if the option could be considered but you can not gather enough information from the case description or the NICE guidelines required to choose this option.

ANSWER RULES:

- Use ONLY information explicitly stated in the transcript

- If the MDT explicitly rejects a standard step (e.g., “no staging required”), preserve that decision and do not reintroduce it.

- Do NOT invent biomarkers, imaging results, or pathology

- If required information is missing, write: INSUFFICIENT INFORMATION

- Use conditional language: "may be considered", "is typically recommended", "depending on eligibility"

- If the transcript contains biomarker testing or pending results, structure the treatment plan with explicit IF/THEN branches.

- Do NOT default to surgery-first or adjuvant-first pathways unless explicitly supported by the transcript.

INTERNAL REASONING POLICY:

- Perform reasoning internally

- Do not output reasoning

-DO not output immediate steps

- Do not explain decisions

- Only output final labels.

```
OUTPUT REQUIREMENTS:

- DO NOT INCLUDE ANY explanations, reasoning, chain-of-thought in the response.

- Return lables only.

==============================================

EXAMPLE:

!---------------Example not provided for data privacy------------!

"""
```

## Supplementary Section 6: Intervention Taxonomy for Structured Prediction

The taxonomy below was supplied to the model in full for every case. The model classified each leaf item as Yes, No or NEI.

1. Surgical Procedures:

1.1. Diagnostic & Image-Guided Procedures

1.1.1. Biopsies (core, vacuum, excisional, punch)

1.1.2. Localisation techniques (wire, seed, magnetic, ultrasound-guided)

1.2. Breast-Conserving Surgery (BCS)

1.2.1. Lumpectomy / Wide Local Excision

1.2.2. Margin re-excision

1.2.3. Oncoplastic Breast-Conserving Surgery

1.3. Mastectomy

1.3.1. Simple, skin-sparing, nipple-sparing

1.3.2. Modified radical

1.3.3. Skin-reducing mastectomy

1.4. Risk-Reducing & Prophylactic Surgery

1.4.1. Bilateral risk-reducing mastectomy

1.4.2. Contralateral prophylactic mastectomy

1.5. Axillary Staging, treatment

1.5.1. Sentinel lymph node biopsy (SLNB)

1.5.2. Axillary lymph node dissection (ALND)

1.5.3. Targeted axillary procedures

1.6. Breast Reconstruction (Oncoplastic & Post-Mastectomy)

1.6.1. Implant-based reconstruction

1.6.2. Autologous flaps (DIEP, LD)

1.6.3. Immediate & delayed reconstruction

1.7. Nipple/Areola Reconstruction

1.7.1. Nipple reconstruction (local flaps, grafts)

1.7.2. Areola tattooing

1.8. Recurrent & Palliative Breast Surgery

1.8.1. Chest wall resection

1.8.2. Surgery for recurrence

1.8.3. Palliative procedures

2. Imaging

2.1. Mammography

2.2. Contrast-Enhanced Mammography (CEM)

2.3. Breast Ultrasound (US)

2.4. Breast MRI

2.5. FDG Positron Emission Tomography-CT (PET-CT)

2.6. Sodium Fluoride PET-CT

2.7. CT Scan of Chest/Abdomen/Pelvis

2.8. Bone Scan (Scintigraphy)

3. Radiotherapy

3.1. Adjuvant Whole Breast Irradiation/ Chest Wall Irradiation (post-mastectomy radiotherapy, PMRT)

3.2. Tumour Bed Boost

3.3. Partial Breast Irradiation (PBI / APBI)

3.4. Regional Nodal Irradiation (RNI)

3.4.1. Axillary irradiation

3.4.2. Supraclavicular nodal irradiation

3.4.3. Internal mammary node (IMN) irradiation

3.5. Stereotactic & Special Techniques- Stereotactic body radiotherapy (SBRT) for oligometastases

3.6. Palliative Radiotherapy

3.6.1. Palliative breast/chest wall irradiation

3.6.2. Radiotherapy for bone metastasis

3.6.3. Radiotherapy to brain metastasis (WBRT / SRS)

3.7. No need for radiotherapy

4. Systemic treatment

4.1. Chemotherapy- Neoadjuvant/adjuvant/palliative chemotherapy

4.2. Endocrine (Hormonal) Therapy

4.2.1. Anti-estrogen therapy- Tamoxifen, Aromatase inhibitors

4.2.2. Ovarian suppression/ablation

4.3. Targeted Therapy (Biologic Therapy)

4.3.1. HER2-targeted therapy

4.3.2. CDK4/6 inhibitors

4.3.3. PI3K / mTOR inhibitors

4.3.4. Antibody-drug conjugates (ADCs)

4.3.5. PARP inhibitors (For BRCA-positive patients)

4.3.6. NTRK targeted (NTRK fusion positive patients)

4.4. Immunotherapy

4.5. Bone-Targeted Therapy- Bisphosphonates, RANKL inhibitors (denosumab)

4.6. Electrochemotherapy

4.7. Supportive treatments

5. Useful Tools

5.1. Prognostic & Treatment Benefit Calculators

5.1.1. PREDICT tool

5.1.2. Nottingham Prognostic Index (NPI)

5.2. Biomarkers

5.2.1. ER / PR status

5.2.2. HER2 status

5.2.3. Ki-67 proliferation index

5.2.4. PD-L1 expression

5.3. Genomic / Molecular Assays- Oncotype Dx

5.4. Common terms used- TNM staging, Tumour grade (Nottingham grading system), Nodal status

6. Referrals

6.1. Referral to other MDT

6.2. Refer to Oncology Health

6.3. Refer for clinical trials

7. Appointments

7.1. Clinical Oncology appointment

7.2. Medical Oncology appointment

# Supplementary Section 7: LLM Prediction Expert Evaluation Instrument and Full Statistical Comparison

## 7.1 Expert Evaluation Instrument (Free-text Generation)

**Step 1. Identify the interventions**

1. Identify all the interventions listed in the MDT GT
2. Identify all the interventions proposed by each LLM

**Step 2. Compare the interventions**

For each intervention proposed by the LLM, answer the following:

Q1. Does the LLM Intervention match the intervention in the MDT GT? (Yes / No)

Q2. If the intervention does not match the MDT plan:

Is it a recognised, real clinical intervention (or made up / hallucinated)?

Q3. If the intervention does not match the MDT plan:

Could it still be clinically appropriate for this specific case?

Q4. Are **ALL** MDT GT interventions present in the LLM response? (Yes / No)

If 'No': list the interventions omitted

Metric definitions, where *Relevant* denotes interventions discussed in the MDT and *Retrieved* denotes interventions generated by the model:

- $\text{Precision} = \frac{|\text{Relevant} \cap \text{Retrieved}|}{|\text{Retrieved}|}$
- $\text{Recall} = \frac{|\text{Relevant} \cap \text{Retrieved}|}{|\text{Relevant}|}$
- $\text{Jaccard Coeff} = \frac{|\text{Relevant} \cap \text{Retrieved}|}{|\text{Relevant} \cup \text{Retrieved}|}$
- $\text{Overgeneration} = \frac{\text{FP}}{|\text{Retrieved}|}$ (from Q1)
- $\text{Miss rate} = \frac{|\text{Missed Interventions}|}{|\text{Relevant}|}$ (from Q4)
- $\text{Hallucination rate} = \frac{|\text{Hallucinated Interventions}|}{|\text{Retrieved}|}$ (from Q2)
- $\text{Proportion Appropriate (proportion of apt interventions)} = \frac{|\text{nonMDT but appropriate}|}{|\text{clinically appropriate interventions}|}$ (from Q3)

### 7.2 Full Paired Statistical Comparison (Intervention-based Strategy)

McNemar's test (MN) was used for overall accuracy and paired bootstrap resampling (B = 5,000; BS) for class-specific precision, recall, and F1 score. Differences are reported as MedGemma-RAG (MR) minus ChatGPT-5.2 (CR); n = 396 paired predictions. Accuracy, precision, recall, and F1 score are reported using standard machine learning definitions.

**Supplementary Table 9** Paired statistical comparison of MedGemma-RAG (MR) and ChatGPT-5.2 (CR) for intervention-based structured prediction under three NEI scoring schemes, reporting overall accuracy and class-specific performance with bootstrap confidence intervals and significance testing.

| Scenario | Class | Metric | MR | CR | Diff | 95% CI | p | Test | Sig. |
|---|---|---|---|---|---|---|---|---|---|
| A (NEI=No) | overall | Accuracy | 83.6% | 80.8% | +2.8 pp | — | 0.1093 | MN | No |
| B (NEI=Yes) | overall | Accuracy | 60.1% | 61.4% | −1.3 pp | — | 0.6301 | MN | No |
| 3-class | overall | Accuracy | 54.0% | 52.5% | +1.5 pp | — | 0.5611 | MN | No |
| A (NEI=No) | Yes | Precision | 0.286 | 0.136 | +0.149 | 0.031–0.275 | 0.0156 | BS | Yes |
| A (NEI=No) | Yes | Recall | 0.318 | 0.136 | +0.182 | 0.041–0.327 | 0.0200 | BS | Yes |
| A (NEI=No) | Yes | F1 | 0.301 | 0.136 | +0.165 | 0.038–0.292 | 0.0136 | BS | Yes |
| A (NEI=No) | No | Precision | 0.914 | 0.892 | +0.021 | 0.004–0.040 | 0.0136 | BS | Yes |
| A (NEI=No) | No | Recall | 0.901 | 0.892 | +0.009 | −0.020–0.037 | 0.6120 | BS | No |
| A (NEI=No) | No | F1 | 0.907 | 0.892 | +0.015 | −0.003–0.033 | 0.0972 | BS | No |
| B (NEI=Yes) | Yes | Precision | 0.228 | 0.245 | −0.017 | −0.047–0.012 | 0.2336 | BS | No |
| B (NEI=Yes) | Yes | Recall | 0.830 | 0.906 | −0.075 | −0.180–0.023 | 0.1964 | BS | No |
| B (NEI=Yes) | Yes | F1 | 0.358 | 0.386 | −0.028 | −0.072–0.016 | 0.1964 | BS | No |
| B (NEI=Yes) | No | Precision | 0.956 | 0.975 | −0.019 | −0.047–0.006 | 0.1496 | BS | No |
| B (NEI=Yes) | No | Recall | 0.566 | 0.569 | −0.003 | −0.049–0.041 | 0.9384 | BS | No |
| B (NEI=Yes) | No | F1 | 0.711 | 0.718 | −0.008 | −0.046–0.030 | 0.6588 | BS | No |
| 3-class | Yes | Precision | 0.286 | 0.136 | +0.149 | 0.031–0.275 | 0.0156 | BS | Yes |
| 3-class | Yes | Recall | 0.318 | 0.136 | +0.182 | 0.041–0.327 | 0.0200 | BS | Yes |
| 3-class | Yes | F1 | 0.301 | 0.136 | +0.165 | 0.038–0.292 | 0.0136 | BS | Yes |
| 3-class | No | Precision | 0.956 | 0.975 | −0.019 | −0.047–0.006 | 0.1496 | BS | No |
| 3-class | No | Recall | 0.566 | 0.569 | −0.003 | −0.049–0.041 | 0.9384 | BS | No |
| 3-class | No | F1 | 0.711 | 0.718 | −0.008 | −0.046–0.030 | 0.6588 | BS | No |
| 3-class | NEI | Precision | 0.042 | 0.046 | −0.004 | −0.020–0.007 | 0.6532 | BS | No |
| 3-class | NEI | Recall | 0.667 | 0.778 | −0.111 | −0.364–0.000 | 0.7400 | BS | No |
| 3-class | NEI | F1 | 0.078 | 0.087 | −0.009 | −0.038–0.011 | 0.6420 | BS | No |